\documentclass{article} % For LaTeX2e
\usepackage{iclr2019_conference,times}
\iclrfinalcopy
\usepackage{amsmath,amsfonts,bm}

\def\eqref#1{equation~\ref{#1}}
\def\1{\bm{1}}

\DeclareMathAlphabet{\mathsfit}{\encodingdefault}{\sfdefault}{m}{sl}
\SetMathAlphabet{\mathsfit}{bold}{\encodingdefault}{\sfdefault}{bx}{n}

\usepackage{hyperref}
\usepackage{url}
\usepackage{fancyhdr,graphicx,amsmath,amssymb}
\usepackage[ruled,vlined]{algorithm2e}
\usepackage{graphicx}
\usepackage{booktabs} 
\usepackage{amsmath}
\usepackage[utf8]{inputenc}
\usepackage[english]{babel}
\usepackage{multicol}
\usepackage[lofdepth,lotdepth]{subfig}

\title{Fast Exploration with Simplified Models and Approximately Optimistic Planning in Model-Based Reinforcement Learning}

\author{Ramtin Keramati\thanks{These authors contributed equally to this work}, Jay Whang\footnotemark[1], Patrick Cho\footnotemark[1], Emma Brunskill \\
Department of Computer Science\\
Stanford University\\
Stanford, CA 94305, USA \\
\texttt{\{keramati,jaywhang,patcho,ebrun\}@cs.stanford.edu} \\
}

\begin{document}

\maketitle
\begin{abstract}

Humans learn to play video games significantly faster than the state-of-the-art reinforcement learning (RL) algorithms. People seem to build simple models that are easy to learn 
%\danote{Learning the model is fast or learning, given the model, is faster?} 
to support planning and strategic exploration. Inspired by this, we investigate two issues in leveraging model-based RL for sample efficiency. First we investigate 
%\danote{If this is a contribution of the paper, I'm guessing you did more than just consider it but actually *show* it. Also, I tend to write in present tense, your call.} 
how to perform strategic exploration when exact planning is not feasible and empirically show that optimistic Monte Carlo Tree Search outperforms posterior sampling methods. Second we show how to learn simple deterministic models to support fast learning using object representation.
%\danote{Again, the OO representation makes the model learning faster or makes learning with the model faster?} 
We illustrate the benefit of these ideas by introducing a novel algorithm, Strategic Object Oriented Reinforcement Learning (SOORL), that outperforms state-of-the-art algorithms in the game of \textit{Pitfall!} in less than 50 episodes.

\end{abstract}

\section{Introduction}
The coupling of deep neural networks and reinforcement learning has led to exciting advances, enabling reinforcement learning agents that can reach human-level performance in many Atari2600 games \citep{mnih2015human, mnih2016asynchronous, silver2017mastering, hessel2017rainbow}.
%\danote{Consider adding maybe and AlphaGo or Rainbow citation so people don't think you stopped looking in 2016}.
However, such agents typically require hundreds of millions of time steps to learn to play well. This is in sharp contrast to people, who typically learn to play Atari games within a few episodes~\citep{lake2017building}. Prior work on human learning for Atari suggests that people may be systematically building models of the reward and dynamics in the domain and using those to plan efficiently \citep{tsividis2017human, dubey2018investigating}.

Given that human learners seemingly perform model-based RL very quickly, we are motivated to consider alternative approaches to current model-based RL, which often involves building complex predictive deep neural networks from scratch. Deep neural networks can require a large amount of data to accurately train, and performing perfect planning with those models is computationally expensive. Indeed, while there is a considerable amount of theoretical work done on tabular model-based reinforcement learning that suggests model-based approaches can be provably sample efficient~\citep{dann2017unifying, brafman2002r, strehl2008analysis}, there is much less success in using model-based reinforcement learning in extremely large environments, in part due to the challenges of learning simple accurate models in these domains. 

Instead, in this paper, we investigate two issues in leveraging model-based RL to speed learning in large domains. First we explore how to perform approximate planning using models during model-based RL to support  exploration, assuming models are given.
%\danote{There's a disconnect between here and the points raised in the previous paragraphs}. 
Models can facilitate deep exploration, since one key benefit of using models is that it is often easier to quantify uncertainty in those models, and perform planning in a way to guide the agent towards exploring and reducing that uncertainty. Recent work has suggested the benefit of Thompson Sampling methods over optimism methods for reinforcement learning~\citep{osband2016posterior}, and, indeed, empirically Thompson Sampling methods have done very well in contextual bandits and small state space MDPs where it is possible to plan exactly. However, to our knowledge, there has not been an investigation of how these popular approaches scale to larger domains where it is computationally prohibitive to perform exact planning. We introduce optimistic MCTS, a variant of the popular planning technique Monte Carlo Tree Search, %\danote{I wouldn't diminish your contribution by calling it small; maybe slight or just get rid of the modifier.}
and find that optimism here outperforms other approaches in several simulation experiments when planning can only be done approximately.

%\danote{I would switch the order of this and the previous paragraph. The paragraph before the previous one talked about difficulties in learning good models; connect your solution to that problem first (this paragraph) and then talking about how that solution also gives way to improved exploration.} 
Second, while our first investigation explores how to plan to encourage exploration given a model, we further investigate how to choose the model class to support computational tractability and learning efficiency. Here we propose to learn object-oriented deterministic models of the domain. People may leverage and test models of object interactions during video game learning~\citep{tsividis2017human, dubey2018investigating}, and object-oriented learning has the nice benefit that data from all similar objects can be pooled when building a model. Prior work has shown that object-oriented model-based reinforcement learning can yield provably efficient learning and scale to larger domains~\citep{diuk2008object}. Deterministic models offer an additional benefit over object-oriented models--they require less data to train (since one does not have to model a large stochastic distribution of outcomes) and they have additional benefits for planning, reducing the branching factor of next states to 1. Indeed past work~\citep{diuk2008object} also proposed learning deterministic object-oriented models. In contrast, we investigate how to learn the transition model and temporal abstraction to make the world appear deterministic. In other words, we assume a candidate set of possible transition model classes and temporal abstractions, and perform model selection to select a level of temporal abstraction and model that deterministically predicts the outcome of an action. %\danote{After reading, 'make the world appear deterministic' still hasn't landed...clarify?}

We illustrate the potential benefit of these ideas by introducing an object oriented algorithm that uses prior knowledge of model classes and object representation, and show that our algorithm can learn to achieve positive reward in the notoriously difficult Atari game \textit{Pitfall!} within 50 episodes. Almost no RL methods have achieved positive reward on \textit{Pitfall!} without human demonstrations, and even with demonstrations, such approaches often take hundreds of millions of frames to learn~\citep{aytar2018playing, hester2017deep}. In contrast to demonstrations, we assume two forms of prior knowledge--a predefined object representation and a class of potential model features. Computer vision is rapidly advancing to the state that soon we will be able to easily extract objects from even artificial scenes like the Arcade Learning Environment.
%\danote{This sounds a bit defensive for this early in the paper. Maybe gently rephrase or consider removing the 'we argue that' and 'that it is quite reasonable'}.
The second assumption of candidate model classes is a stronger assumption, but a large set of models could be defined directly given the object classes and only incur a cost quadratic in the set of features. %\rknote{Why this is true? :D}

While encouraging, such results should be mostly viewed as a case study. We believe the key contributions of our paper are not this particular demonstration on \textit{Pitfall!}, but rather the investigation of exploration approaches when planning can only be done approximately, and the benefit of selecting among model representations to support computationally tractable planning and fast learning. The second can be viewed as a bias/variance tradeoff and in future work we plan to consider how to identify and account for model biases that could limit asymptotic performance during the model-based planning procedure.

\section{Preliminaries}

We consider a finite horizon Markov Decision Process (MDP) $\langle \mathcal{S}, \mathcal{A}, T, R, \gamma \rangle$, where $\mathcal{S}$ is the state space, $\mathcal{A}$ the action space, $T:\mathcal{S} \times \mathcal{A} \rightarrow \mathcal{S}$ the transition function, $R:\mathcal{S} \times \mathcal{A} \rightarrow \mathbb{R}$ the reward function, and $\gamma$ the discount factor. The goal of the RL agent is to maximize the expected discounted reward $\mathbb{E}_{\pi}[\sum_{t=0}^{T}\gamma^t R(s_t, a_t)]$ following a policy $\pi$. 

\section{Approximate Model-Based Planning to Support Exploration}\label{sec:exp}

Computing an optimal policy for a Markov decision process in a large state space can be extremely computationally expensive, if not impossible. Therefore some form of approximate planning procedure is almost always needed for large scale  domains. Especially in model-based RL algorithms when the transition and reward models are initially unknown and should be learned online %\danote{online RL $\implies$ model-based RL?}
, a computationally tractable planning algorithm is more critical. Since the model estimates are changing and one should plan more frequently with the new model estimates.
% particularly critical in online RL methods where the model estimates are changing -- must be computationally tractable

One popular and powerful approach for scaling up planning for large Markov decision processes is Monte Carlo Tree Search~\citep{chaslot2008monte, browne2012survey}, which has been used to achieve better than world class performance in Go~\citep{silver2016mastering}. When a perfect transition and reward model is known, MCTS is guaranteed to converge to the optimal value function in the limit of infinite computation (i.e. infinite number of rollouts). However, when a model is estimated from the data and computational power is bounded, prior works has suggested how to adjust planning horizon to get the best computational performance~\citep{jiang2015dependence}. As of our knowledge there is no work that additionally investigate the performance of different exploration methods with approximate planning.%\rknote{--rewrite: this paragraph doesn't motivate the next one}
% MCTS is guaranteed to eventually converge to the optimal value given infinite computation and perfect dynamics and reward models
% Given bounded amount of computation, prior research has suggested how to adjust the horizon used in order to yield the best computational performance (NAN)

We are interested in how to leverage models to support deep efficient exploration in large domains. Prior research has strong guarantees to find a near optimal policy with deep exploration when exact planning can be done, often in small state space~\citep{brunskill2009provably, brafman2002r}. In contrast, in this work, we investigate how to leverage models to support deep exploration when exact planning is not possible (large state space). 
% in most theoretical work, state space small, planning done exactly
% other times assume feasible to do e-optimally (cite my JMLR 2009 paper)
% in contrast, want to consider how to leverage models to support exploration when can't plan exactly

We propose a new optimistic MCTS to support deep exploration guided by models. Particularly, we suggest to perform a MCTS algorithm (e.g. UCT~\citep{kocsis2006bandit}), and use the learned reward $R$ and transition model $T$ that is optimistic toward unseen or less frequently seen part of the state-action space (we further discuss how to learn simple models for planning in section \ref{sec:model}). 
%\danote{A little more detail might help really nail down what differentiates your modified MCTS from regular MCTS} 
Concretely, while performing rollouts using learned models at each step an optimistic reward bonus is given. Optimistic reward bonus can be given by any optimism in the face of uncertainty (OFU) algorithm (e.g. MBIE-EB~\citep{strehl2008analysis} or Rmax~\citep{brafman2002r}). %\rknote{pseudo code might be useful}

Optimistic MCTS may have some advantages over alternate ways to achieve deep exploration in large state spaces given limited computation. 
In particular optimistic MCTS is leveraging model uncertainty to drive deep exploration, in contrast to policy search methods with simulated models~\citep{sutton2000policy} that relies on the stochasticity of the policy for exploration and it is unclear how to leverage model uncertainty.

Additionally, MCTS with posterior sampling methods, like Thompson Sampling~\citep{thompson1933likelihood} (by sampling a model from posterior distribution and performing rollouts) has strong guarantees when exact planning can be done, might not be optimistic enough for efficient exploration. With limited number of rollouts, agent might not observe the optimistic part of the model, in contrast to optimistic MCTS where optimism is built into every node of the tree.%\rknote{BACMP is more specific that TS, I put the discussion about BAMCP after experiments with more details}  

% policy search with simulated models -- unclear how to leverage model uncertainty to encourage deep exploration
% Thompson sampling (sample a model and run MCTS). Problem is that proofs that TS work normally rely on solving exactly
%     and that some parts of the world may be optimistically estimated. With limited rollouts, may not reach those 
%     optimistic estimates
% BAMCP: samples many models. approximating POMDP but larger branching factors

To investigate the impact of approximate planning on deep exploration, we compare optimistic MCTS, with Thomson Sampling, and BAMCP~\citep{guez2012efficient} (a tractable sample-based method for
approximate Bayes-optimal planning) in two toy domains, \textit{Pong Prime} and \textit{mini-Pitfall!}.

\begin{table}[tb]
  \begin{center}
    \begin{tabular}{r|c|c|c}
      \toprule 
      \textbf{Method} & Optimism & Thompson Sampling & BAMCP\\
      \midrule
      \textbf{Number of Episdoes} & 5.0 $\pm$ 1.7 & 6.2 $\pm$ 1.95 & 11.0 $\pm$ 2.5\\       
      \bottomrule % <-- Midrule here
      \end{tabular}
      
  \end{center}
  \caption{Comparison of different exploration methods in \textit{mini-Pitfall!} to consistently achieve the reward at the right end of the first room.}
  \label{tab:mini_pitfall}
\end{table}

\begin{figure}[tb]
    \centering
    \subfloat[Pong Prime environment]{\includegraphics[width=.31\linewidth]{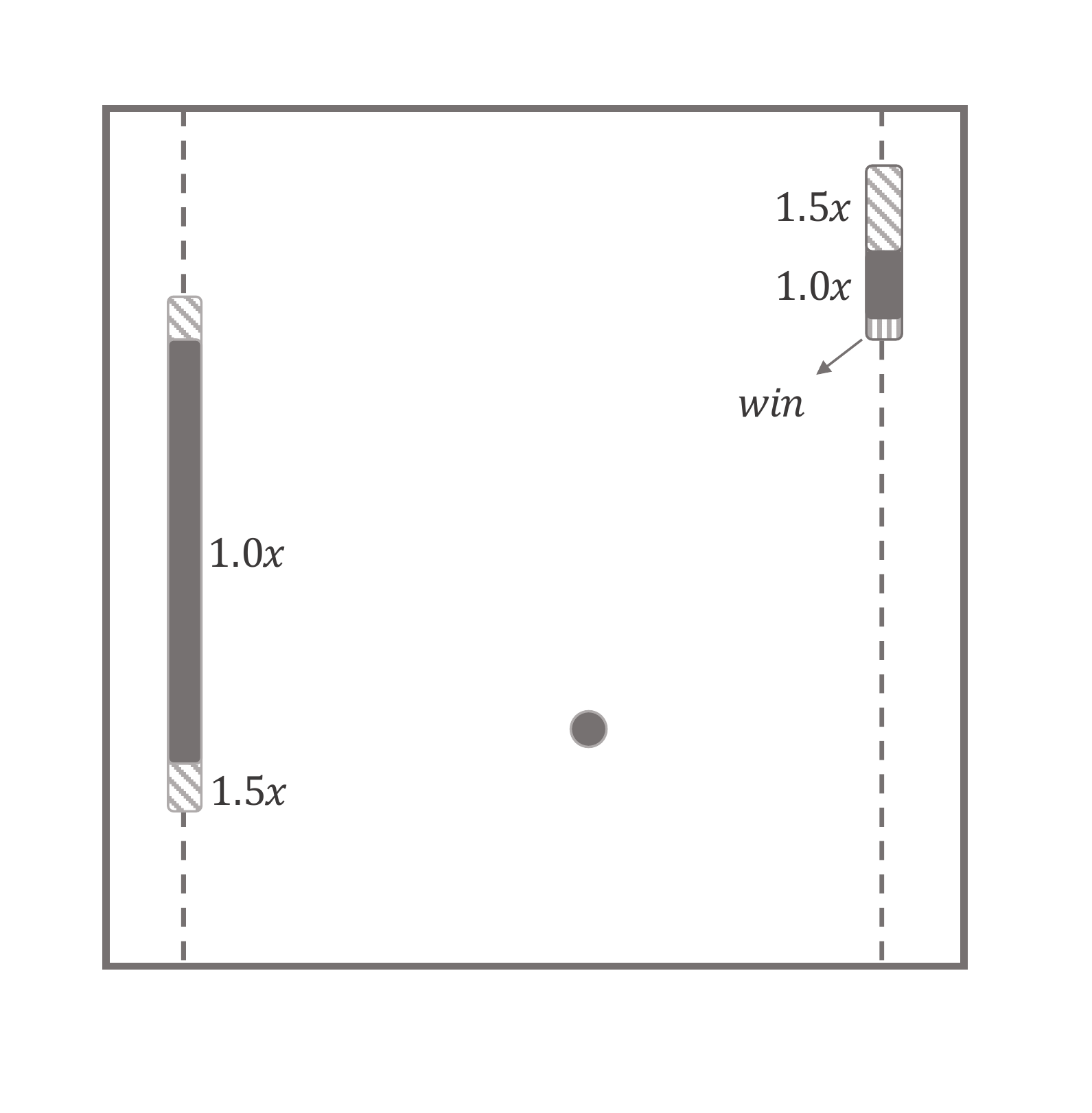}}%
    % \enskip
    \subfloat[Different exploration methods]{\includegraphics[width=.33\linewidth]{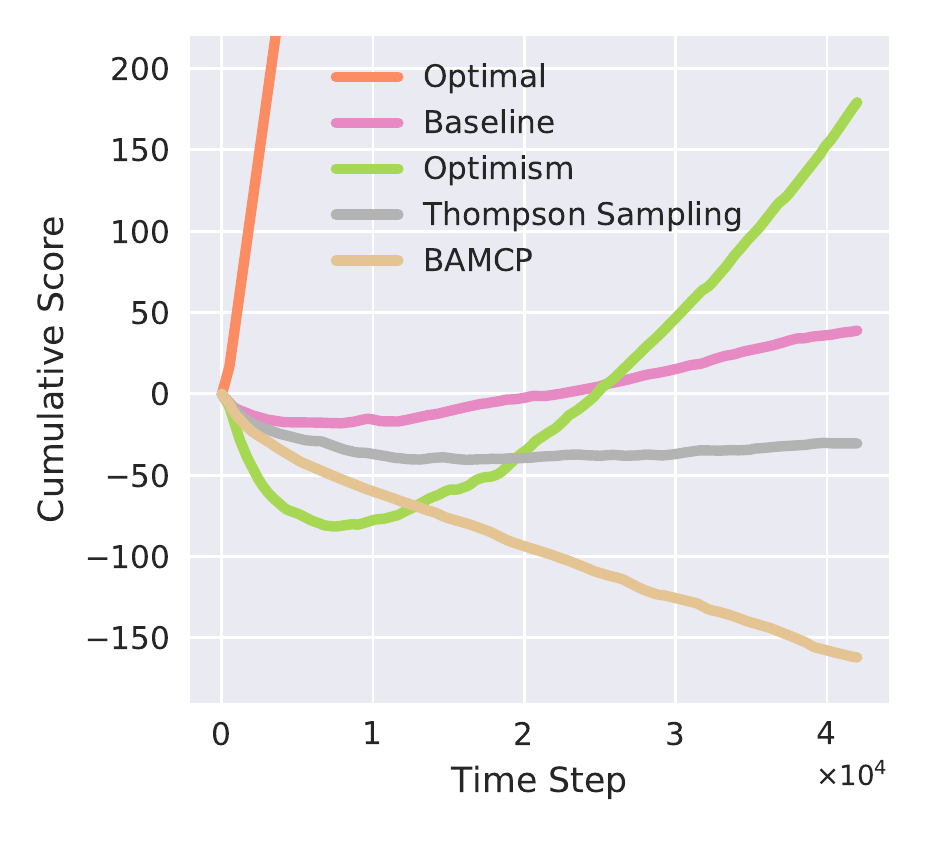}}%
    % \enskip
    \subfloat[Optimism based exploration]{\includegraphics[width=.33\linewidth]{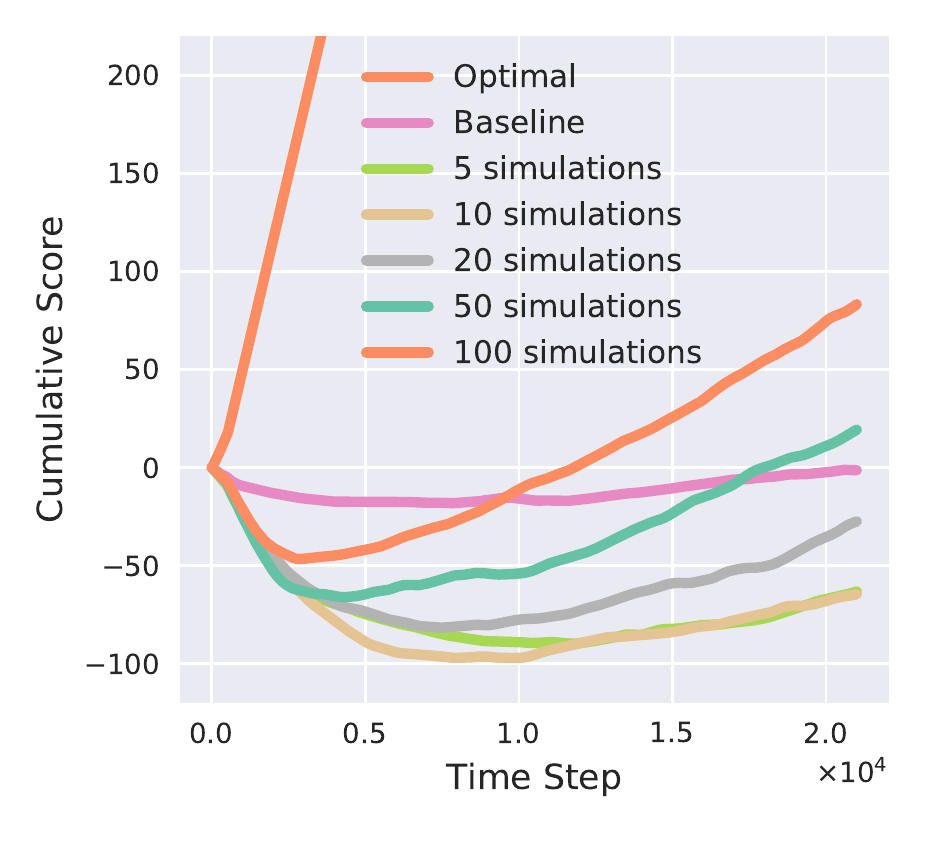}}%
    \caption{Pong Prime}%
    \label{fig:pongp}%
\end{figure}

\subsection{Experiments}

% \eb{Discuss why we use these two to investigate the impact of imperfect model-based planning for deep exploration. E.g. wanted to choose small scale environments where exploration is needed, similar in spirit to general atari domains. }

In order to test our hypotheses on the impact of imperfect model-based planning on deep exploration we introduced two toy environments, \textit{PongPrime} and \textit{mini-Pitfall!} that are similar to general Atari games but in smaller scale. \textit{Pong Prime} environment is designed for a hard exploration task. Dynamics of this game is similar to Atari2600's Pong environment~\citep{bellemare2013arcade} with minor tweaks that make the game significantly harder. The enemy paddle is made 3 times larger than the player paddle.
%so it is impossible to score a point by simply hitting the ball back at the normal speed.
Additionally, the top and bottom 10\% percent of the enemy paddle hit the ball back at 1.5 times the normal speed.
%so that the enemy paddle is even more powerful. 
Similarly, the player paddle also consists of 3 regions with distinct behavior. The \emph{top region} of the paddle hits the ball back at 1.5 times speed. The \emph{middle region} hits the ball back at normal speed. Finally, the \emph{lower region} covers 5\% of the paddle and instantly wins a point for the player. This configuration is set up so that it is difficult but not impossible for the player to score using the top region (scoring on average around 5\% of the time the ball bounces off the top region). In this setting, the optimal policy is to always hit the ball with the lower region of the ball. The game is deterministic and model free methods with $\epsilon$-greedy exploration (e.g. DQN) consistently loses the game with lowest possible score across 5000 episodes. Figure\ref{fig:pongp}(a) shows this environment. 

\textit{mini-Pitfall!} is a small version of Atari2600 game \textit{Pitfall!} which we use as a final test bed of our algorithm. In this version we limit the agent to two rooms of the game (the initial room, room 0, that agents starts the game in, and the room on the left side, room -1). There exist a dummy reward $R_{max}$ at the right end of the room 0, and the left end of room -1 is a terminal state (underground connection is also terminal state). Figure \ref{fig:meta_action}(a) shows this environment. 

We provided the right model class for both experiments so we can separate the effect of exploration from model mismatch. Figure \ref{fig:pongp}(b) compares the performance of different exploration strategies to the baseline (which uses a MLE model with UCT algorithm) combining with UCT algorithm in \textit{Pong Prime} domain. We perform 500 total tree searches for all runs in Figure \ref{fig:pongp}(b). Additionally, Table \ref{tab:mini_pitfall} shows the performance of different exploration methods in achieving the reward on the right side of the room 0 in \textit{mini-Pitfall!} consistently.

\subsection{Discussion}\label{sec:exp_disc}

Both BAMCP and TS perform worse than the MLE model. In the limit of infinite simulations, BAMCP is guaranteed to converge to the Bayes optimal solution~\citep{guez2012efficient}. Similarly, with full horizon planning, we should be able to compute the exact value for the model sampled with TS, and there are strong guarantees that such a method will converge to the optimal policy~\citep{osband2016posterior}. 
%However in practice, especially in large domains or domains with real-time constraints, the amount of computation, and therefore the quality of the computed plan, will be significantly limited. 
However, if it is infeasible to use a depth that mimics the game horizon, or perhaps even to reach a local reward, then TS may suffer. This is because TS samples a single model, which means that parts of the model may be overly optimistic, while other parts may be pessimistic. Hence, when performing a limited number of simulations using UCT, we may not go down branches of the tree that \textit{observe} the optimistic parts of the sampled model. Therefore, the computed estimates of the Q value at the root node may not be optimistic, which is often a key part of proofs of the effectiveness of TS methods, and very helpful empirically. %\rknote{This paragraph has a lot of overlap with hypothesis we made}

Additionally, BAMCP suffers more in these environments that are deterministic. This means that for TS, optimism, and MLE approaches, the tree constructed will only have one child node (the deterministic next state) for any chosen action. In contrast, BAMCP samples a different deterministic model at each rollout, and for the same action node, those models may each deterministically predict different next states. Hence, BAMCP with $M$ sampled models and planning horizon $H$, potentially builds a tree of size O($(|A|M)^H$), in contrast to the other methods that build a tree of at most size O($|A|^H$), where $|A|$ is the size of action space.

Optimism-based exploration significantly outperforms other approaches. Optimism is built into \textit{every} node of the three that is allowing it to distinguish even locally between actions that may need exploration, in absence of observing long delayed reward. As we demonstrate in Figure \ref{fig:pongp}(c) for the optimistic method, as planning power increases through more simulations (number of rollouts), the performance of optimism-based exploration also increases. With sufficient computational power, optimistic MCTS should learn the optimal policy for \textit{Pong Prime} domain.

\section{Approximately deterministic model to support fast learning}\label{sec:model}

In the previous section we considered how to plan and perform deep exploration given a transition and reward models. Now we ask a natural question, what types of model to learn?, and how these models will affect computational tractability of planning and sample efficient learning? In general simpler models are easier to learn and requires less data to train, in contrast to complex function approximation methods that often requires massive amount of data and suffer from compounding errors in lookahead planning~\citep{roderick2017deep, weber2017imagination}. Additionally, simple models are tractable to perform deep exploration with model uncertainty and reduce planning time.

There are reasonable evidence that humans learns a simple model, often inaccurate, to support planning and guide decision making~\citep{tsividis2017human, dubey2018investigating}. In particular, people seem to benefit from higher level object representations that allows them to factor a high dimensional state space into simple low dimensional object states, this allows human to generalize from few examples, explore and plan efficiently. 
Additionally, as discussed in section \ref{sec:exp_disc}, deterministic models can significantly help in planning and deep exploration, due to smaller branching factor of the tree (each state node will have only one child per action). Inspired by humans and prior works on object oriented MDP~\citep{diuk2008object} we hypothesize that object oriented models can help us learn a simple, easy and approximately deterministic models that are sufficient for planning .

Object detection has been long studied in computer vision, and state-of-the-art algorithms can detect objects with great accuracy in real world scenes (e.g. YoLo~\citep{redmon2017yolo9000}, fast RCNN~\citep{girshick2015fast} mask RCNN~\citep{he2017mask} and \dots). We expect that these algorithms can simply detect objects and their bounding boxes, when they are trained on Atari2600. Thus, here we assume objects are given, in section \ref{sec:model_learning} we discuss how to learn a simple model and further in section \ref{sec:macro} we show how temporal abstraction can help learning an approximately deterministic mode.

% Learning a transition and reward models for large state MDPs often requires massive amount of data and the use of complex function approximations methods, that usually suffers from compounding errors~\citep{roderick2017deep, weber2017imagination} and are not well suited for lookahead planning methods like MCTS. For a sample efficient model-based RL algorithm, learning a simple model (that requires minimum amount of data) that can support planning (i.e. doesn't suffer from compounding errors) is vital. Additionally, performing deep exploration with complex function is often intractable and not computationally feasible.\rknote{CITE??}

% Prior works on human learning for Atari~\citep{tsividis2017human, dubey2018investigating} highlights people's ability to generalize from few examples, explore and plan efficiently. Additionally, people seem to benefit from higher level object representations that allows them to factor a high dimensional state space into simple low dimensional object states. Inspired by humans and prior works on object oriented MDP~\citep{diuk2008object} we hypothesize that object oriented models can help us learn a simple and easy models that are sufficient for planning.\rknote{relation to factored MDP and finding a set of parent features}

\subsection{Review of Object Oriented MDP}

We use a simpler version of OOMDP~\citep{diuk2008object}. we define a set of object classes $\mathcal{C} = \{c_1, \dots, c_n\}$ where each class has a set of attributes $\{c.a_1,\dots,c.a_m\}$. Each state $s$ consists of objects $f(s) = \{o_1,\dots,o_k\}$ where each object $o_i \in \mathcal{C}$. The state of an object is defined by the value assignment to its attributes. Finally, the state  $s$ of the underlying MDP is the union of all object states $\cup_{i=1}^{k} o_i$. 

We define the interaction function $I: \mathcal{O} \times \mathcal{O} \rightarrow \{0,1\}$ to be an indicator that determines if two objects are interacting with each other. For simplicity, we make three assumptions: first, that this interaction function is known; second, objects from the same class share the same transition function; and third, each object's next state is  dependent on at most pairwise object interactions and action.
An object's successor state is determined by a standalone transition function $T_c(o, a)$ or a pairwise transition function $T_{c_i,c_j}(o_i,o_j,a)$ if $I(o_i,o_j) = 1$.

\begin{figure}[tb]
    \centering
    \subfloat[mini \textit{Pitfall!}]{\includegraphics[width=.25\linewidth]{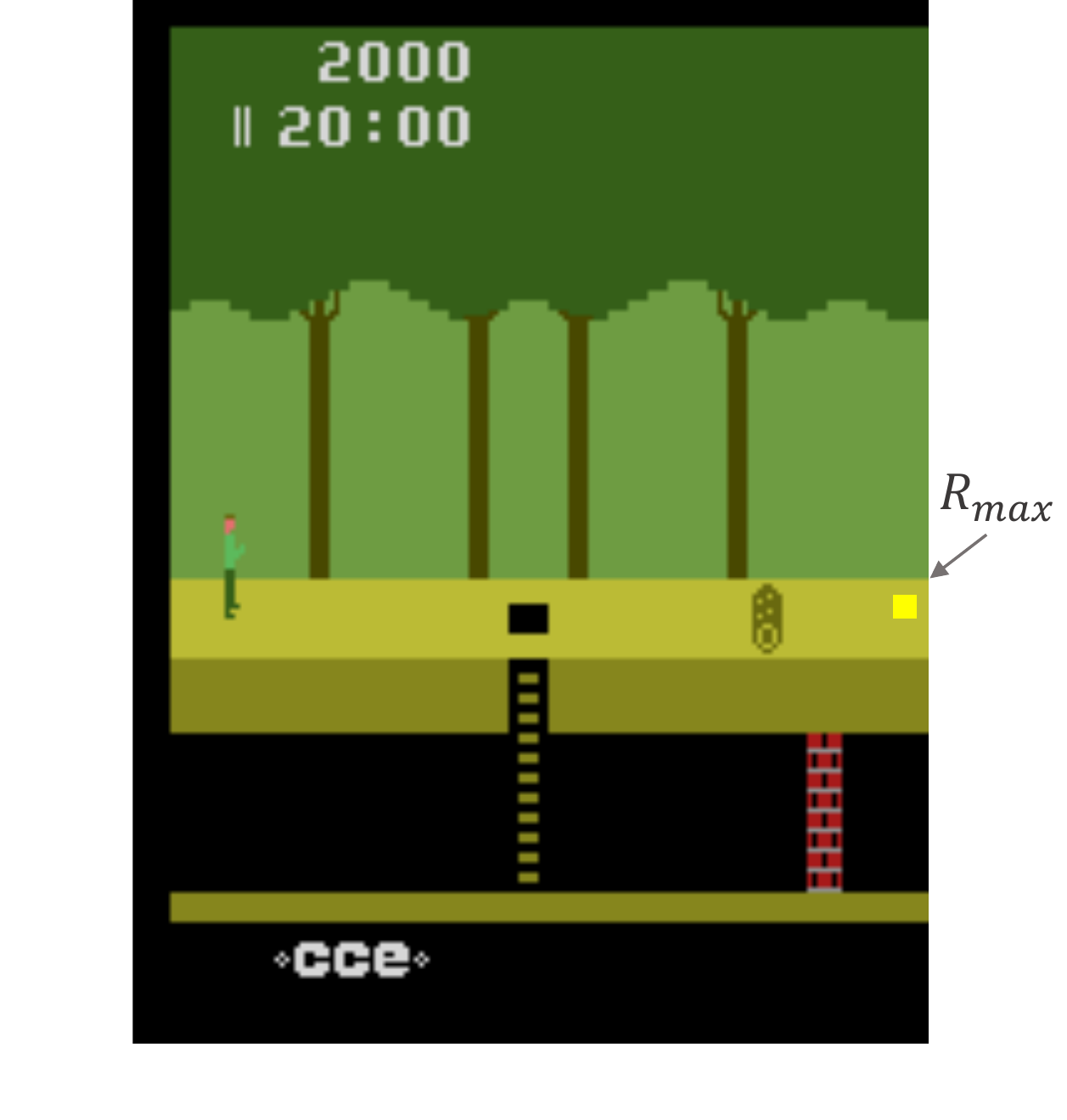}}%
    % \enskip
    \subfloat[][Total Entropy of Models]{\includegraphics[width=.37\linewidth]{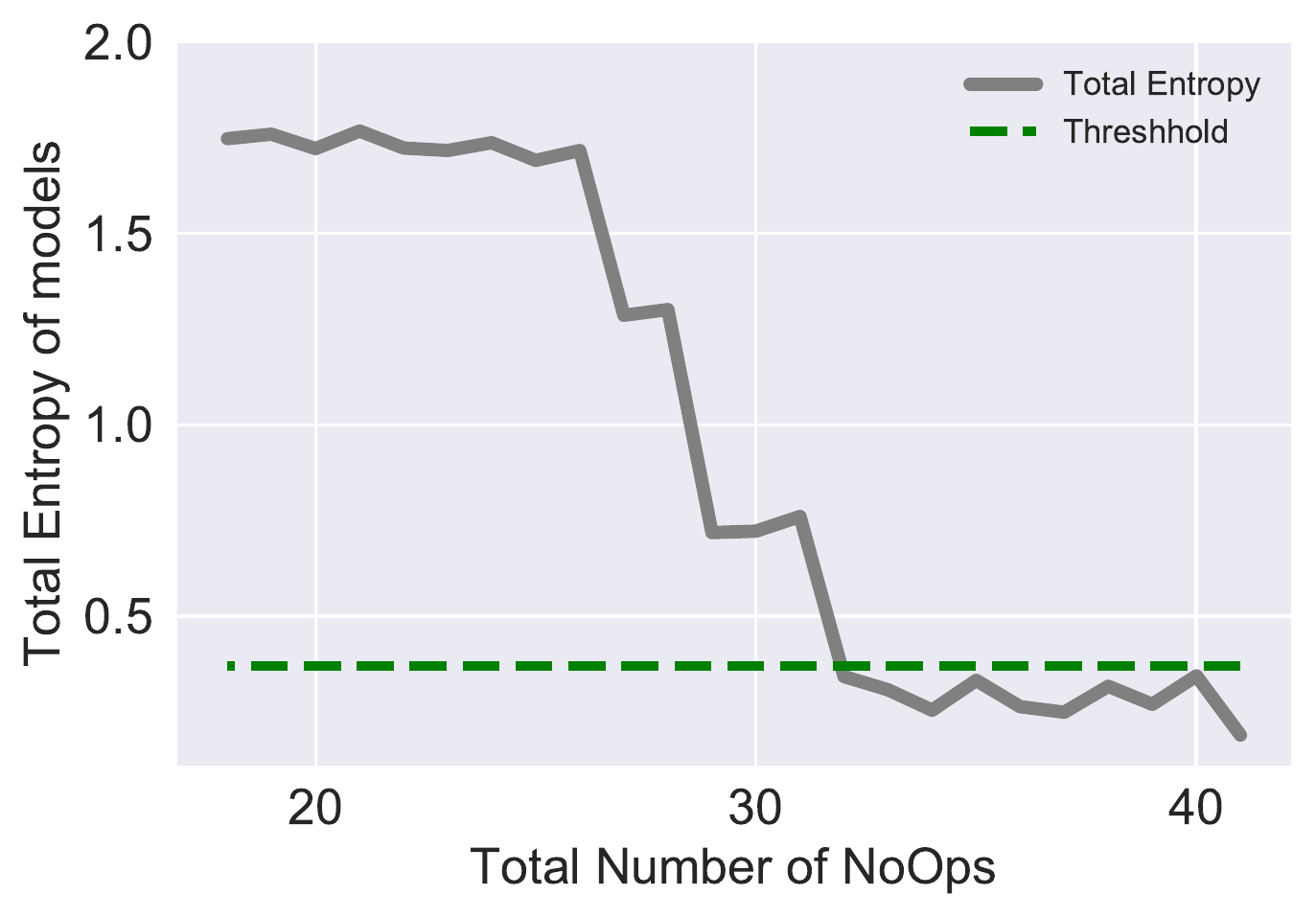}}%
    % \enskip
    \subfloat[][Per Action Entropy of Agent's Model]{\includegraphics[width=.37\linewidth]{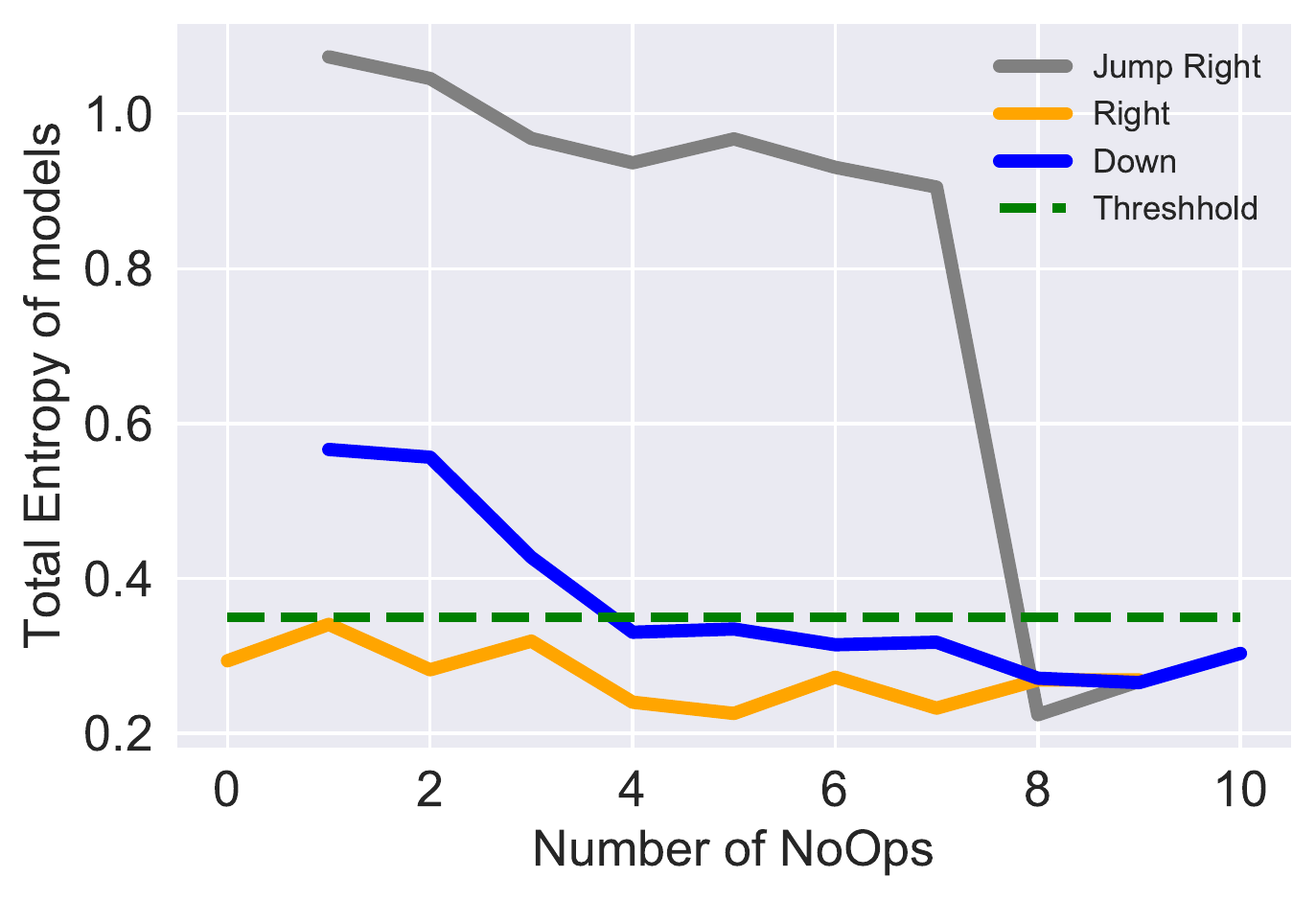}}%
    \caption{Macro actions}%
    \label{fig:meta_action}%
\end{figure}

\subsection{Model Learning}\label{sec:model_learning}

Given that we are planning at an object level, we hypothesize that even simple models, such as linear and discrete count based models, give sufficient accuracy for planning, since often objects follow a very simple physical laws. More importantly, to ensure "sufficient accuracy in planning", we further require that these models predict transitions and rewards in a deterministic fashion.

To ensure deterministic transitions, we consider the class of functions $\mathcal{F}_t = \{f_{t}^{1},\dots,f_{t}^{n}\}$, where each $f_{t}^{i}$ is a count-based model of the dynamics for an object. Each function stores the count of every output based on a different set of input features with given history $t$. The simplest model in $\mathcal{F}_1$ is $f_{1}^{1}$, which uses one history with null input. For example, for a falling object with steady state velocity, such a model is sufficient as we can predict displacement $\delta x$ and $\delta y$ without any input. On the other hand, $f_{1}^{n}$, which uses one history and the most complex set of features, is the most complicated model in the class $\mathcal{F}_1$. In terms of objects, the most complex set of input features that we consider is the union of the object's state features, relative state features with respect to an interacting object, and action. 

The goal then is to choose the simplest model that achieves deterministic transitions within $\mathcal{F}_t$. To do so, we compute the entropy of the observed data for each function $H(f_{t}^{k}) = -\sum_{x_i} p(x_i) \log(p(x_i|f_{t}^{k}))$ where the summation is over all the observed data. We choose the simplest model that has entropy less than a predefined threshold $\epsilon_{ent}$. If none of the models in $\mathcal{F}_t$ satisfy the entropy threshold, we increase $t$ through an exponential back off scheme. Concretely, we increase the history to the next exponent of 2. We use the same approach and same class of models for reward functions.

\begin{figure}[tb]
    \centering

    \subfloat[Model entropy]{\includegraphics[width=.36\linewidth]{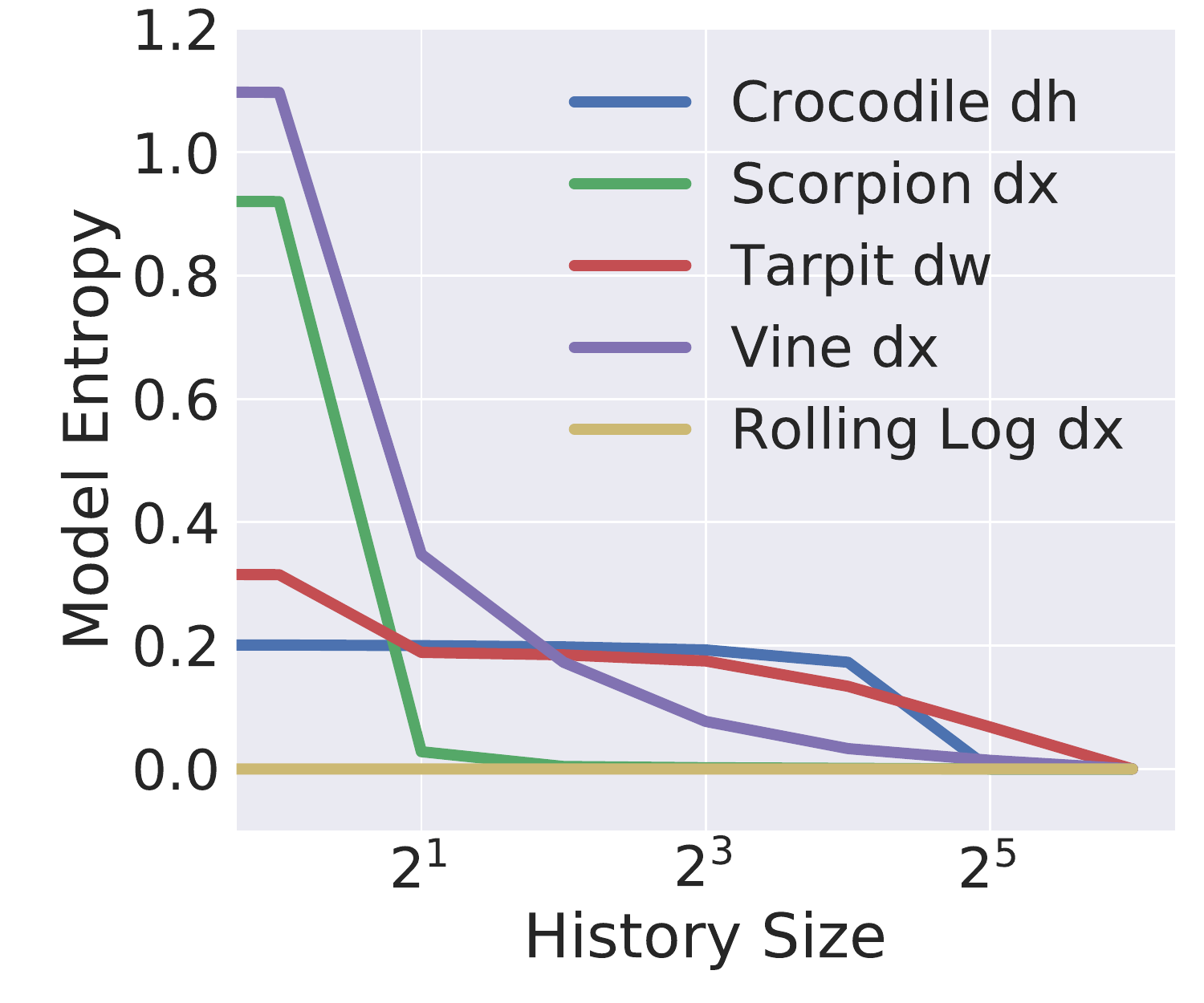}}%
    \quad
    \subfloat[Example trained models]{\includegraphics[width=.595\linewidth]{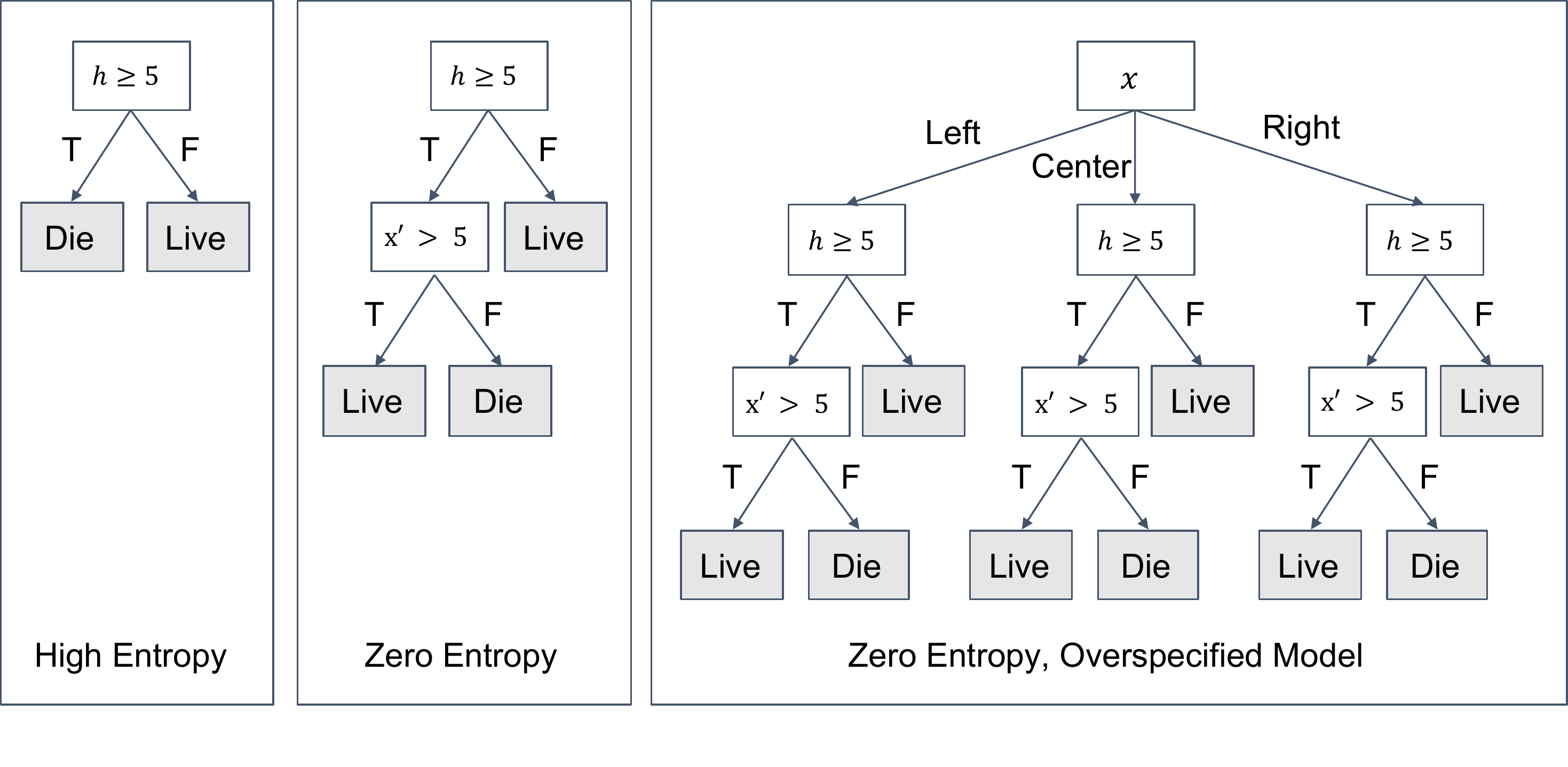}}%
    \quad
    \caption{Model selection}%
    
    \label{fig:pitfall:metaction}%
\end{figure}

Figure \ref{fig:pitfall:metaction}(a) shows an example for the game \textit{Pitfall!}, where we used a Cartesian product of object size $(w,h)$, object location $(x,y)$ and object intersection $(x',y')$ as features. Ignoring null input, this Cartesian product results in 7 different feature sets. With sufficient history, the entropy of all the models eventually drops to zero. 

\subsection{Temporal Abstraction}\label{sec:macro}

As we observed in our experiments, objects transitions, especially action-dependent transitions, can show a highly nonlinear behaviour and dependency to multiple time step histories. Inspired by human's \textit{reaction time} and previous work~\citep{diuk2008object} we use the notion of macro action in the form of \textit{"act and then wait"} in order to learn a simple approximately deterministic transition models. 

Algorithm \ref{alg:macro} shows the pseudo code to learn the macro actions from a predefined set of atomic actions (e.g. in the simplest form it can be the action space of the desired MDP; however, one can define them as any n-token combination of actions). In algorithm \ref{alg:macro} we augment all atomic actions with $k$ number of \textit{no-op} or wait, and then greedily decrease the number of no-ops followed by each action such that all model's can deterministically predict the next object state. The  goal is to have a model that achieves deterministic transition, to do so we measure the entropy of model's prediction with the observed data as in section \ref{sec:model_learning}.

\begin{algorithm}
\KwIn{maximum number of no-op $k$, set of atomic actions $\mathcal{A}$}
\SetAlgoLined
macro actions $\gets$ atomic action followed by maximum number of no-op\;
mark all macro action as reducible\;
\While{there exist a reducible macro action}{
    $a \gets$select a reducible macro action and decrease number of no-op by one\;
    $\tau \gets$ compute entropy of model's prediction with new set of macro actions\;
    \If{$\tau \geq thresh$}{
        mark $a$ as non reducible macro action and restore the number of no-op for $a$\;
    }
}
 \caption{\bf Macro Actions}\label{alg:macro}
\end{algorithm}

Figure \ref{fig:meta_action}(b) shows the total entropy of model's versus total number of \textit{no-op} in \textit{Pitfall!} environment. As algorithm progress entropy increases to pass the threshold. Figure \ref{fig:meta_action} (c) shows the entropy of agent's models while reducing number of \textit{no-op} for one action and keeping others constant. As it shows Jump Right requires 8, Down requires 4 and Right requires none \textit{no-op} afterward to achieves a deterministic transition for all actions. Running this algorithm in \textit{Pong Prime} environment results in two \textit{no-op} after each action, since the real dynamics uses 3 step history.

\section{SOORL: Strategic Object Oriented Reinforcement Learning}
In this section we put together the insights from previous sections and propose a novel model-based object oriented RL algorithm, Strategic Object Oriented RL (\textbf{SOORL}). SOORL assumes access to an object detector, that returns a list of objects with their attributes (i.e. location and bounding box), an interaction function and macro actions (that can be learned with algorithm \ref{alg:macro}).

Algorithm \ref{alg:soorl} shows a pseudo code of SOORL. At each step, SOORL performs lookahead planning with UCT algorithm, learn and select appropriate transition and reward models for object representation and performs optimism based exploration. 

\begin{algorithm}
\KwIn{object detector $f(s)$, lookahead planning depth $d$, number of rollouts $l$}
initialize\;
\For{each episode e}{
    train value function V on replay buffer $\mathcal{D}$\;
    \For{each step i}{
    $o_i \gets$ detect objects with object detector $f(s_i)$\;
    $Q(s_i,a) \gets$ perform lookahead planning with depth $d$ and $l$ rollouts\;
    take action $a_i ={argmax}{Q(s,.)}$ and update replay buffer $\mathcal{D} = \mathcal{D} \cup (o_i, a_i, o_{i+1}, r_i)$\;
    update transition and reward models with $(o_i, a_i, o_{i+1}, r_i)$
    }
}
\caption{{\bf SOORL} \label{alg:soorl}}
\end{algorithm}

\textbf{Model Learning:} SOORL uses the method described in section \ref{sec:model_learning} with Cartesian product of object size $(w,h)$, object location $(x,y)$ and object intersection $(x',y')$ as features. Ignoring null input, this Cartesian product results in 7 different feature sets.

\textbf{Exploration:} Count based models allow SOORL to efficiently perform the knows what it knows (KWIK)~\citep{li2008knows} scheme for exploration (optimistic MCTS). Concretely, if our algorithm queries the transition or reward model with a previously unseen input, we consider the resulting state as a state with $R_{max}$ reward. $R_{max}$ reward is also considered for any previously unseen object interactions. As we observe the reward for each interaction we update the reward model based on model-based interval estimation~\citep{strehl2008analysis}.

\textbf{Planning:} At the beginning of each episode, a value function is trained based on previously seen transitions and rewards. Value function $V: \mathcal{O} \rightarrow \mathbb{R}$ is trained over object states and can generally be any function approximation methods.

Lookahead planning is performed by UCT~\citep{kocsis2006bandit} algorithm with $l$ rollouts and depth $d$. Algorithm \ref{alg:planning} shows pseudo code of planning, at each planning step, SOORL computes object interactions, selects appropriate models based on interactions and object states then predicts rewards and next object state. At depth $d$ of planning SOORL uses value of the object state $V(o_d)$ trained at the beginning of each episode.

\begin{algorithm}[h]
\KwIn{objects $o$, depth $d$, rollouts $l$, value function $V$}
\For{number of rollouts $l$}{
    set current depth $i$ to zero\;
    \While{$i \leq d$}{
        \If{$i == d$}{
            update $Q$ values with $V(o_i)$\;
            break\;
        }
        detect interactions $I_i$ with objects $o_i$ and select models $T,R$\;
        take action $a_i = argmax Q(s_i,a) + c \sqrt{\frac{log N(s,a)}{N(s)}}$\;
        obtain next object state $o_{i+1}$ and reward $r_{i}$ with $R(o_i,I_i), T(o_i,I_i)$ and update $Q$\;
    }
}
\caption{{\bf Lookahead Planning} \label{alg:planning}}
\end{algorithm}

\section{Empirical Evaluation}\label{sec:pitfall}

In this section we will evaluate SOORL on an Atari game \textit{Pitfall!}. SOORL assumes access to an object detector, a predefined set of function classes and macro action (that can be learned using algorithm described in section \ref{sec:macro}). 

Labeled data for object's in Atari game in not available, and in our experiment we extracted the objects from Atari RAM and screen information. The need for an object detector makes engineering burden of SOORL prohibitive to test the algorithm on all Atari games. Thus we focused on one of the hardest game (sparse reward and hard exploration~\citep{bellemare2016unifying}), \textit{Pitfall!}, where all the previous methods (without human demonstrations) failed to achieve any positive reward. We showed that object representation can be extremely helpful in this hard exploration game and SOORL can achieve a positive reward in \textit{Pitfall!} without human demonstration. 

\textit{Pitfall!} is an Atari2600 environment where the goal is to have the agent traverse through multiple rooms (255 in total) while collecting rewards and avoiding obstacles. It is arguably one of the hardest Atari2600 game~\citep{hester2017deep} due to its large map, sparse positive and dense negative rewards that necessitate deep exploration and long-horizon planning. The $\epsilon$-greedy exploration strategy completely fails in this environment, and more recent count-based exploration \citep{bellemare2016unifying} does not show much performance boost due to the sparsity of positive reward. Pitfall is difficult even for human players without prior knowledge of the game -- \citep{hester2017deep} reports that human performance varies from 3662 to 47821 points, whereas for other hard Atari games, this variation is much smaller (e.g. from 32300 to 34900 for Montezuma's revenge).

\subsection{Details}
All objects information are extracted from RAM and screen information, and each object's attribute is location $(x,y)$ and bounding box size $(w,h)$. Objects are considered interacting with each other if bounding boxes collide. Transition and reward models for each object are based on the method described in Section \ref{sec:model_learning}. The features used are a Cartesian product of object size $(w,h)$, object location $(x,y)$ and object intersection $(x',y')$. Ignoring null input, this Cartesian product results in 7 different feature sets.

As described in section \ref{sec:exp} we used optimism based exploration method, by assigning reward $R_{max}$ to all unseen interactions and transitions. As we observe reward for each interaction we update the model, based on model based interval estimation~\citep{strehl2008analysis}. Additionally in order to further incentivize exploration we split the screen into $N \times  M$ grids and keeping a count of the number of times agent visits each grid. The agent is given a reward bonus $\beta n(s)^{-1/2}$ based on visit count $n(s)$.

For value function, we used discrete object's location with the same split used for optimism bonus, and computed the empirical transition $\hat{T}(o'|o,a)$ and empirical reward with optimism bonus $\hat{r}(o,a) + \beta n(s)^{-1/2}$, and at the beginning of each episode, we performed value iteration.

\subsection{Performance and Discussion}

% \textbf{**EB: Can we also try plotting episodes vs number of times that run got a positive reward in an episode (so mostly 0 then 1 then 2 ...) and then average these? Add add error bars}\rknote{I didn't understand how to plot this}

\begin{figure}[tb]
    \centering
    \subfloat[Room Discovery]{\includegraphics[width=.29\linewidth]{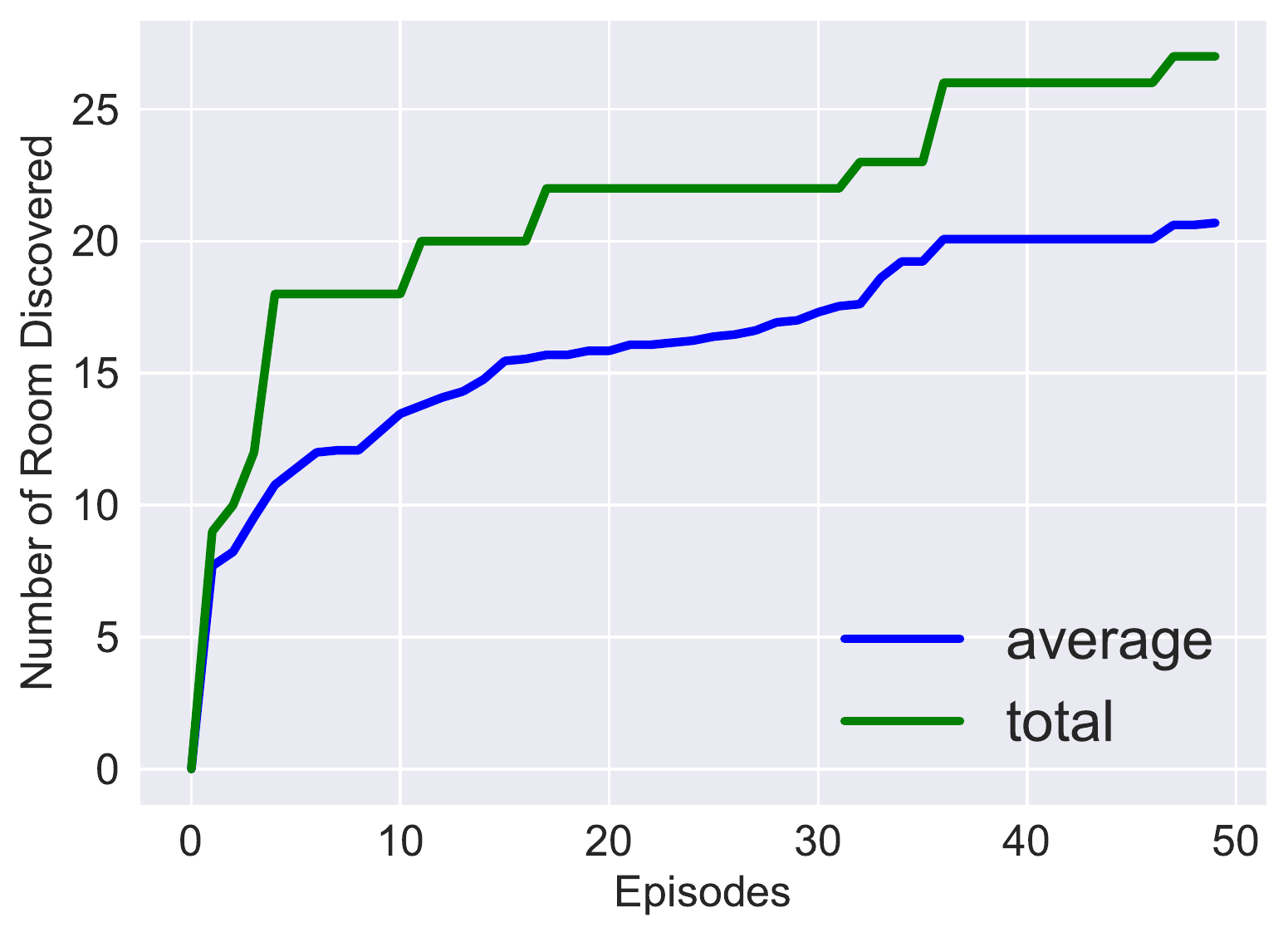}}%
    \subfloat[][Reward per episode]{\includegraphics[width=.31\linewidth]{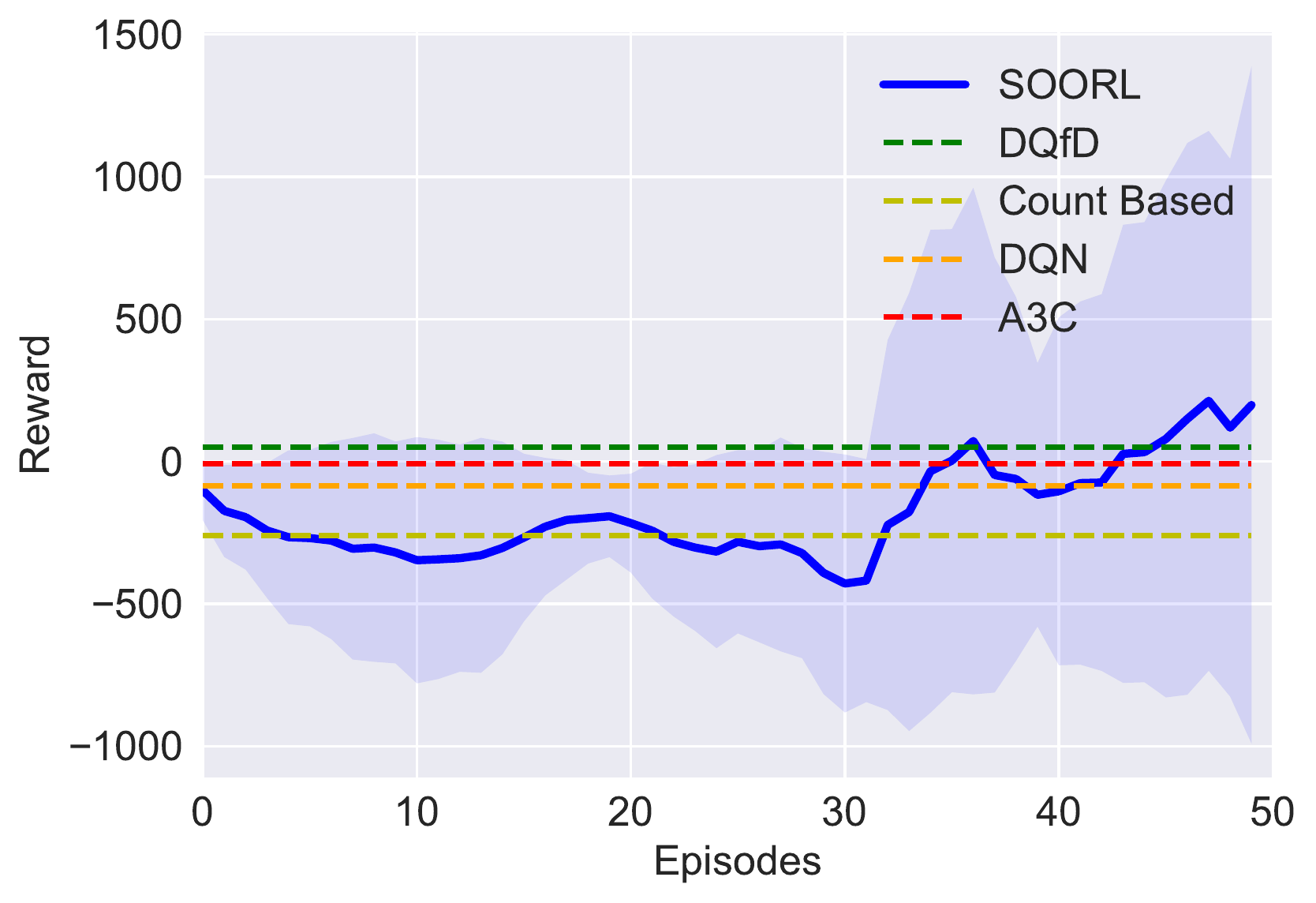}}%
    \subfloat[][Percentage of runs with positive reward]{\includegraphics[width=.33\linewidth]{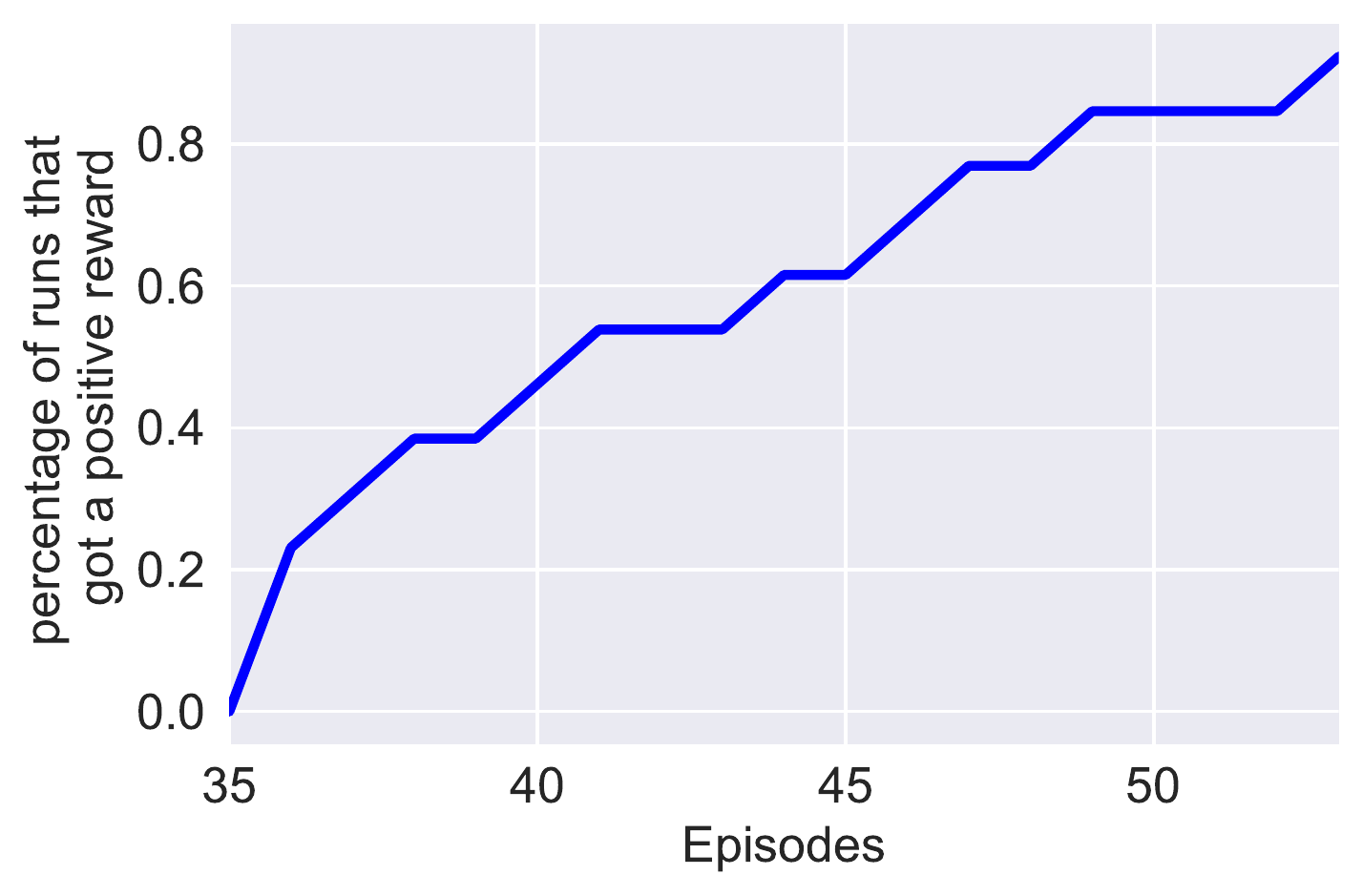}}%
    \caption{performance of SOORL on Pitfall!}%
    \label{fig:pitfall}%
\end{figure}

Figure \ref{fig:pitfall}(a) shows an increasing number of rooms being discovered across episodes. On average, the agent discovers 21 rooms within 50 episodes, and in total across 20 runs the agent discovers 27 different rooms. This validates our hypothesis that optimistic MCTS drives deep exploration, that we also showed in a smaller domains in section \ref{sec:exp}. 

 Figure \ref{fig:pitfall}(b) shows accumulative reward for each episode of SOORL and compares them to the other state of the art algorithm. Results of count-based \citep{bellemare2016unifying}, DQfD \citep{hester2017deep}, A3C \citep{mnih2016asynchronous} and DQN \citep{mnih2015human} are reported at the time of evaluation. Our average score across all episodes and all runs is $-193.5 \pm 595.8$ and SOORL score for the best episode across all runs in $606.6 \pm 1254.5$, which is higher than all scores that were reported at the time of evaluation. To the best of our knowledge, this is the first approach which manages to get positive rewards on \textit{Pitfall!} without human demonstrations.  Sample videos of the agent reaching the two closest positive rewards can be found here: \url{https://youtu.be/GvenPZMJiTg} (4000 reward)\,\,
\url{https://youtu.be/74F-ta5LyuA} (2000 reward)

Figure \ref{fig:pitfall} (c) shows the percentage of runs that got a positive reward. More than $80\%$ of runs got a positive reward in 50 episodes that shows consistency of our approach across multiple runs. Due to simple dynamics model that can be learned fast SOORL is extremely sample efficient in comparison to other deep RL method that often takes millions of frame to find a good policy. However, end-to-end deep RL methods use significantly less prior knowledge and using raw pixels as input, thus a direct comparison with them is unfair. We have also provided macro actions and object information for DQN and DDQN but those methods are not designed to take advantage of object representation and did not show a boost in performance. On the other hand, SOORL uses much less prior knowledge than methods with human demonstration~\citep{aytar2018playing, hester2017deep}, where human guidance enormously reduce the challenge of exploration.

Additionally, macro actions are provided for SOORL, but as shown in section \ref{sec:macro} these can be learned online. Integrating this temporal abstraction with SOORL can increase the sample complexity by $\mathcal{O}(|A|K)$, where $K$ is the maximum number of no-ops.

\section{Related Work}

Model based RL has accomplished great success in tasks in which a perfect simulator is known~\citep{silver2016mastering, silver2017mastering}, mostly using Monte Carlo Tree Search algorithms like UCT~\citep{kocsis2006bandit, chaslot2010monte}. Recent work has focused on applying deep learning to model-based RL by learning the model online~\citep{rosin2011nested, finn2017deep, lenz2015deepmpc, weber2017imagination}. In contrast to these methods, we seek to learn a simple model based on object representation.

Exploration has been extensively studied in the tabular setting~\citep{brafman2002r,strehl2008analysis,jaksch2010near, osband2016posterior}. However, these methods do not scale well to large MDPs and often result in poor sample complexity. Additionally, these methods assume exact planning that might not be feasible in large state space. Recent approaches~\citep{bellemare2016unifying, ostrovski2017count, tang2017exploration} proposed an extension of count based exploration to large MDPs. Despite good asymptotic performance, when compared with humans \citep{tsividis2017human, lake2017building}, these methods require orders of magnitude more samples.

Bayesian RL methods also provide an effective balance between exploration and exploitation~\citep{ghavamzadeh2015bayesian}. When exact planning is possible, they provide strong guarantees. However, these methods remain computationally intractable in large state spaces. Recent works propose an extension of these methods to large MDPs~\citep{osband2016deep, azizzadenesheli2018efficient, fortunato2017noisy}. However, these methods are unable to show substantial improvement in hard exploration, sparse reward environments. 

A closely related line of research, finite horizon planning~\citep{kearns2002sparse, mannor2007bias, kearns1994introduction}, has noted how planning horizon can affect planning loss~\citep{kearns2002near, strehl2009reinforcement}. Recent study has shown that shorter planning horizons might be better when there is model inaccuracy~\citep{jiang2015dependence}. In contrast, in this work, we studied how imperfect planning can affect exploration.

OOMDP \citep{diuk2008object} defines a notion borrowed from relational MDPs \citep{guestrin2003generalizing}, and uses objects to learn models and perform model based planning. Model free methods that use object representation \citep{garnelo2016towards, roderick2017deep, cobo2013object} fail to scale to large MDPs and do not leverage object representation for deep exploration. The main difference between our approach and other object oriented approaches is that we perform scalable planning with strategic exploration by leveraging objects to learn simple dynamics models.

\section{Conclusion and Future Work}
To conclude, we showed how we can achieve a sample efficient RL algorithm with object priors. We proposed optimistic MCTS as a way to drive deep exploration when exact planning is impossible, and showed this to be more effective than posterior sampling methods. Additionally, we investigate how approximately deterministic simple models can be learned with object representation to support fast learning and planning.

We introduced Strategic Object Oriented RL (SOORL) that uses object representation and optimistic MCTS with automatic model selection that biases towards simple deterministic models. SOORL achieves state of the art results in the game of $\textit{Pitfall!}$. While there remains works to be done to reduce the engineering burden of SOORL, lookahead planning with object representation is a very promising path towards more sample efficient RL algorithms.

A very important line of future research is \textbf{robust planning}. One important challenge in model-based RL is making planning robust to model
inaccuracy. Identifying the right model class is a nontrivial task, and a wrong model class can
easily introduce a catastrophic error in long-horizon prediction that prohibits the use of tree search
algorithms like UCT.

\bibliography{iclr2019_conference}

\begin{thebibliography}{50}
\providecommand{\natexlab}[1]{#1}
\providecommand{\url}[1]{\texttt{#1}}
\expandafter\ifx\csname urlstyle\endcsname\relax
  \providecommand{\doi}[1]{doi: #1}\else
  \providecommand{\doi}{doi: \begingroup \urlstyle{rm}\Url}\fi

\bibitem[Aytar et~al.(2018)Aytar, Pfaff, Budden, Paine, Wang, and
  de~Freitas]{aytar2018playing}
Yusuf Aytar, Tobias Pfaff, David Budden, Tom~Le Paine, Ziyu Wang, and Nando
  de~Freitas.
\newblock Playing hard exploration games by watching youtube.
\newblock \emph{arXiv preprint arXiv:1805.11592}, 2018.

\bibitem[Azizzadenesheli et~al.(2018)Azizzadenesheli, Brunskill, and
  Anandkumar]{azizzadenesheli2018efficient}
Kamyar Azizzadenesheli, Emma Brunskill, and Animashree Anandkumar.
\newblock Efficient exploration through bayesian deep q-networks.
\newblock \emph{arXiv preprint arXiv:1802.04412}, 2018.

\bibitem[Bellemare et~al.(2016)Bellemare, Srinivasan, Ostrovski, Schaul,
  Saxton, and Munos]{bellemare2016unifying}
Marc Bellemare, Sriram Srinivasan, Georg Ostrovski, Tom Schaul, David Saxton,
  and Remi Munos.
\newblock Unifying count-based exploration and intrinsic motivation.
\newblock In \emph{Advances in Neural Information Processing Systems}, pp.\
  1471--1479, 2016.

\bibitem[Bellemare et~al.(2013)Bellemare, Naddaf, Veness, and
  Bowling]{bellemare2013arcade}
Marc~G Bellemare, Yavar Naddaf, Joel Veness, and Michael Bowling.
\newblock The arcade learning environment: An evaluation platform for general
  agents.
\newblock \emph{Journal of Artificial Intelligence Research}, 47:\penalty0
  253--279, 2013.

\bibitem[Brafman \& Tennenholtz(2002)Brafman and Tennenholtz]{brafman2002r}
Ronen~I Brafman and Moshe Tennenholtz.
\newblock R-max-a general polynomial time algorithm for near-optimal
  reinforcement learning.
\newblock \emph{Journal of Machine Learning Research}, 3\penalty0
  (Oct):\penalty0 213--231, 2002.

\bibitem[Browne et~al.(2012)Browne, Powley, Whitehouse, Lucas, Cowling,
  Rohlfshagen, Tavener, Perez, Samothrakis, and Colton]{browne2012survey}
Cameron~B Browne, Edward Powley, Daniel Whitehouse, Simon~M Lucas, Peter~I
  Cowling, Philipp Rohlfshagen, Stephen Tavener, Diego Perez, Spyridon
  Samothrakis, and Simon Colton.
\newblock A survey of monte carlo tree search methods.
\newblock \emph{IEEE Transactions on Computational Intelligence and AI in
  games}, 4\penalty0 (1):\penalty0 1--43, 2012.

\bibitem[Brunskill et~al.(2009)Brunskill, Leffler, Li, Littman, and
  Roy]{brunskill2009provably}
Emma Brunskill, Bethany~R Leffler, Lihong Li, Michael~L Littman, and Nicholas
  Roy.
\newblock Provably efficient learning with typed parametric models.
\newblock \emph{Journal of Machine Learning Research}, 10\penalty0
  (Aug):\penalty0 1955--1988, 2009.

\bibitem[Chaslot et~al.(2008)Chaslot, Bakkes, Szita, and
  Spronck]{chaslot2008monte}
Guillaume Chaslot, Sander Bakkes, Istvan Szita, and Pieter Spronck.
\newblock Monte-carlo tree search: A new framework for game ai.
\newblock In \emph{AIIDE}, 2008.

\bibitem[Chaslot(2010)]{chaslot2010monte}
Guillaume Maurice Jean-Bernard~Chaslot Chaslot.
\newblock \emph{Monte-carlo tree search}.
\newblock PhD thesis, Maastricht University, 2010.

\bibitem[Cobo et~al.(2013)Cobo, Isbell, and Thomaz]{cobo2013object}
Luis~C Cobo, Charles~L Isbell, and Andrea~L Thomaz.
\newblock Object focused q-learning for autonomous agents.
\newblock In \emph{Proceedings of the 2013 international conference on
  Autonomous agents and multi-agent systems}, pp.\  1061--1068. International
  Foundation for Autonomous Agents and Multiagent Systems, 2013.

\bibitem[Dann et~al.(2017)Dann, Lattimore, and Brunskill]{dann2017unifying}
Christoph Dann, Tor Lattimore, and Emma Brunskill.
\newblock Unifying pac and regret: Uniform pac bounds for episodic
  reinforcement learning.
\newblock In \emph{Advances in Neural Information Processing Systems}, pp.\
  5713--5723, 2017.

\bibitem[Diuk et~al.(2008)Diuk, Cohen, and Littman]{diuk2008object}
Carlos Diuk, Andre Cohen, and Michael~L Littman.
\newblock An object-oriented representation for efficient reinforcement
  learning.
\newblock In \emph{Proceedings of the 25th international conference on Machine
  learning}, pp.\  240--247. ACM, 2008.

\bibitem[Dubey et~al.(2018)Dubey, Agrawal, Pathak, Griffiths, and
  Efros]{dubey2018investigating}
Rachit Dubey, Pulkit Agrawal, Deepak Pathak, Thomas~L Griffiths, and Alexei~A
  Efros.
\newblock Investigating human priors for playing video games.
\newblock \emph{arXiv preprint arXiv:1802.10217}, 2018.

\bibitem[Finn \& Levine(2017)Finn and Levine]{finn2017deep}
Chelsea Finn and Sergey Levine.
\newblock Deep visual foresight for planning robot motion.
\newblock In \emph{Robotics and Automation (ICRA), 2017 IEEE International
  Conference on}, pp.\  2786--2793. IEEE, 2017.

\bibitem[Fortunato et~al.(2017)Fortunato, Azar, Piot, Menick, Osband, Graves,
  Mnih, Munos, Hassabis, Pietquin, et~al.]{fortunato2017noisy}
Meire Fortunato, Mohammad~Gheshlaghi Azar, Bilal Piot, Jacob Menick, Ian
  Osband, Alex Graves, Vlad Mnih, Remi Munos, Demis Hassabis, Olivier Pietquin,
  et~al.
\newblock Noisy networks for exploration.
\newblock \emph{arXiv preprint arXiv:1706.10295}, 2017.

\bibitem[Garnelo et~al.(2016)Garnelo, Arulkumaran, and
  Shanahan]{garnelo2016towards}
Marta Garnelo, Kai Arulkumaran, and Murray Shanahan.
\newblock Towards deep symbolic reinforcement learning.
\newblock \emph{arXiv preprint arXiv:1609.05518}, 2016.

\bibitem[Ghavamzadeh et~al.(2015)Ghavamzadeh, Mannor, Pineau, Tamar,
  et~al.]{ghavamzadeh2015bayesian}
Mohammad Ghavamzadeh, Shie Mannor, Joelle Pineau, Aviv Tamar, et~al.
\newblock Bayesian reinforcement learning: A survey.
\newblock \emph{Foundations and Trends{\textregistered} in Machine Learning},
  8\penalty0 (5-6):\penalty0 359--483, 2015.

\bibitem[Girshick(2015)]{girshick2015fast}
Ross Girshick.
\newblock Fast r-cnn.
\newblock In \emph{Proceedings of the IEEE international conference on computer
  vision}, pp.\  1440--1448, 2015.

\bibitem[Guestrin et~al.(2003)Guestrin, Koller, Gearhart, and
  Kanodia]{guestrin2003generalizing}
Carlos Guestrin, Daphne Koller, Chris Gearhart, and Neal Kanodia.
\newblock Generalizing plans to new environments in relational mdps.
\newblock In \emph{Proceedings of the 18th international joint conference on
  Artificial intelligence}, pp.\  1003--1010. Morgan Kaufmann Publishers Inc.,
  2003.

\bibitem[Guez et~al.(2012)Guez, Silver, and Dayan]{guez2012efficient}
Arthur Guez, David Silver, and Peter Dayan.
\newblock Efficient bayes-adaptive reinforcement learning using sample-based
  search.
\newblock In \emph{Advances in Neural Information Processing Systems}, pp.\
  1025--1033, 2012.

\bibitem[He et~al.(2017)He, Gkioxari, Doll{\'a}r, and Girshick]{he2017mask}
Kaiming He, Georgia Gkioxari, Piotr Doll{\'a}r, and Ross Girshick.
\newblock Mask r-cnn.
\newblock In \emph{Computer Vision (ICCV), 2017 IEEE International Conference
  on}, pp.\  2980--2988. IEEE, 2017.

\bibitem[Hessel et~al.(2017)Hessel, Modayil, Van~Hasselt, Schaul, Ostrovski,
  Dabney, Horgan, Piot, Azar, and Silver]{hessel2017rainbow}
Matteo Hessel, Joseph Modayil, Hado Van~Hasselt, Tom Schaul, Georg Ostrovski,
  Will Dabney, Dan Horgan, Bilal Piot, Mohammad Azar, and David Silver.
\newblock Rainbow: Combining improvements in deep reinforcement learning.
\newblock \emph{arXiv preprint arXiv:1710.02298}, 2017.

\bibitem[Hester et~al.(2017)Hester, Vecerik, Pietquin, Lanctot, Schaul, Piot,
  Horgan, Quan, Sendonaris, Dulac-Arnold, et~al.]{hester2017deep}
Todd Hester, Matej Vecerik, Olivier Pietquin, Marc Lanctot, Tom Schaul, Bilal
  Piot, Dan Horgan, John Quan, Andrew Sendonaris, Gabriel Dulac-Arnold, et~al.
\newblock Deep q-learning from demonstrations.
\newblock \emph{arXiv preprint arXiv:1704.03732}, 2017.

\bibitem[Jaksch et~al.(2010)Jaksch, Ortner, and Auer]{jaksch2010near}
Thomas Jaksch, Ronald Ortner, and Peter Auer.
\newblock Near-optimal regret bounds for reinforcement learning.
\newblock \emph{Journal of Machine Learning Research}, 11\penalty0
  (Apr):\penalty0 1563--1600, 2010.

\bibitem[Jiang et~al.(2015)Jiang, Kulesza, Singh, and
  Lewis]{jiang2015dependence}
Nan Jiang, Alex Kulesza, Satinder Singh, and Richard Lewis.
\newblock The dependence of effective planning horizon on model accuracy.
\newblock In \emph{Proceedings of the 2015 International Conference on
  Autonomous Agents and Multiagent Systems}, pp.\  1181--1189. International
  Foundation for Autonomous Agents and Multiagent Systems, 2015.

\bibitem[Kearns \& Singh(2002)Kearns and Singh]{kearns2002near}
Michael Kearns and Satinder Singh.
\newblock Near-optimal reinforcement learning in polynomial time.
\newblock \emph{Machine learning}, 49\penalty0 (2-3):\penalty0 209--232, 2002.

\bibitem[Kearns et~al.(2002)Kearns, Mansour, and Ng]{kearns2002sparse}
Michael Kearns, Yishay Mansour, and Andrew~Y Ng.
\newblock A sparse sampling algorithm for near-optimal planning in large markov
  decision processes.
\newblock \emph{Machine learning}, 49\penalty0 (2-3):\penalty0 193--208, 2002.

\bibitem[Kearns et~al.(1994)Kearns, Vazirani, and
  Vazirani]{kearns1994introduction}
Michael~J Kearns, Umesh~Virkumar Vazirani, and Umesh Vazirani.
\newblock \emph{An introduction to computational learning theory}.
\newblock MIT press, 1994.

\bibitem[Kocsis \& Szepesv{\'a}ri(2006)Kocsis and
  Szepesv{\'a}ri]{kocsis2006bandit}
Levente Kocsis and Csaba Szepesv{\'a}ri.
\newblock Bandit based monte-carlo planning.
\newblock In \emph{European conference on machine learning}, pp.\  282--293.
  Springer, 2006.

\bibitem[Lake et~al.(2017)Lake, Ullman, Tenenbaum, and
  Gershman]{lake2017building}
Brenden~M Lake, Tomer~D Ullman, Joshua~B Tenenbaum, and Samuel~J Gershman.
\newblock Building machines that learn and think like people.
\newblock \emph{Behavioral and Brain Sciences}, 40, 2017.

\bibitem[Lenz et~al.(2015)Lenz, Knepper, and Saxena]{lenz2015deepmpc}
Ian Lenz, Ross~A Knepper, and Ashutosh Saxena.
\newblock Deepmpc: Learning deep latent features for model predictive control.
\newblock In \emph{Robotics: Science and Systems}, 2015.

\bibitem[Li et~al.(2008)Li, Littman, and Walsh]{li2008knows}
Lihong Li, Michael~L Littman, and Thomas~J Walsh.
\newblock Knows what it knows: a framework for self-aware learning.
\newblock In \emph{Proceedings of the 25th international conference on Machine
  learning}, pp.\  568--575. ACM, 2008.

\bibitem[Mannor et~al.(2007)Mannor, Simester, Sun, and
  Tsitsiklis]{mannor2007bias}
Shie Mannor, Duncan Simester, Peng Sun, and John~N Tsitsiklis.
\newblock Bias and variance approximation in value function estimates.
\newblock \emph{Management Science}, 53\penalty0 (2):\penalty0 308--322, 2007.

\bibitem[Mnih et~al.(2015)Mnih, Kavukcuoglu, Silver, Rusu, Veness, Bellemare,
  Graves, Riedmiller, Fidjeland, Ostrovski, et~al.]{mnih2015human}
Volodymyr Mnih, Koray Kavukcuoglu, David Silver, Andrei~A Rusu, Joel Veness,
  Marc~G Bellemare, Alex Graves, Martin Riedmiller, Andreas~K Fidjeland, Georg
  Ostrovski, et~al.
\newblock Human-level control through deep reinforcement learning.
\newblock \emph{Nature}, 518\penalty0 (7540):\penalty0 529, 2015.

\bibitem[Mnih et~al.(2016)Mnih, Badia, Mirza, Graves, Lillicrap, Harley,
  Silver, and Kavukcuoglu]{mnih2016asynchronous}
Volodymyr Mnih, Adria~Puigdomenech Badia, Mehdi Mirza, Alex Graves, Timothy
  Lillicrap, Tim Harley, David Silver, and Koray Kavukcuoglu.
\newblock Asynchronous methods for deep reinforcement learning.
\newblock In \emph{International Conference on Machine Learning}, pp.\
  1928--1937, 2016.

\bibitem[Osband \& Van~Roy(2016)Osband and Van~Roy]{osband2016posterior}
Ian Osband and Benjamin Van~Roy.
\newblock Why is posterior sampling better than optimism for reinforcement
  learning.
\newblock \emph{arXiv preprint arXiv:1607.00215}, 2016.

\bibitem[Osband et~al.(2016)Osband, Blundell, Pritzel, and
  Van~Roy]{osband2016deep}
Ian Osband, Charles Blundell, Alexander Pritzel, and Benjamin Van~Roy.
\newblock Deep exploration via bootstrapped dqn.
\newblock In \emph{Advances in neural information processing systems}, pp.\
  4026--4034, 2016.

\bibitem[Ostrovski et~al.(2017)Ostrovski, Bellemare, Oord, and
  Munos]{ostrovski2017count}
Georg Ostrovski, Marc~G Bellemare, Aaron van~den Oord, and R{\'e}mi Munos.
\newblock Count-based exploration with neural density models.
\newblock \emph{arXiv preprint arXiv:1703.01310}, 2017.

\bibitem[Redmon \& Farhadi(2017)Redmon and Farhadi]{redmon2017yolo9000}
Joseph Redmon and Ali Farhadi.
\newblock Yolo9000: better, faster, stronger.
\newblock \emph{arXiv preprint}, 2017.

\bibitem[Roderick et~al.(2017)Roderick, Grimm, and Tellex]{roderick2017deep}
Melrose Roderick, Christopher Grimm, and Stefanie Tellex.
\newblock Deep abstract q-networks.
\newblock \emph{arXiv preprint arXiv:1710.00459}, 2017.

\bibitem[Rosin(2011)]{rosin2011nested}
Christopher~D Rosin.
\newblock Nested rollout policy adaptation for monte carlo tree search.
\newblock In \emph{Ijcai}, pp.\  649--654, 2011.

\bibitem[Silver et~al.(2016)Silver, Huang, Maddison, Guez, Sifre, Van
  Den~Driessche, Schrittwieser, Antonoglou, Panneershelvam, Lanctot,
  et~al.]{silver2016mastering}
David Silver, Aja Huang, Chris~J Maddison, Arthur Guez, Laurent Sifre, George
  Van Den~Driessche, Julian Schrittwieser, Ioannis Antonoglou, Veda
  Panneershelvam, Marc Lanctot, et~al.
\newblock Mastering the game of go with deep neural networks and tree search.
\newblock \emph{nature}, 529\penalty0 (7587):\penalty0 484--489, 2016.

\bibitem[Silver et~al.(2017)Silver, Hubert, Schrittwieser, Antonoglou, Lai,
  Guez, Lanctot, Sifre, Kumaran, Graepel, et~al.]{silver2017mastering}
David Silver, Thomas Hubert, Julian Schrittwieser, Ioannis Antonoglou, Matthew
  Lai, Arthur Guez, Marc Lanctot, Laurent Sifre, Dharshan Kumaran, Thore
  Graepel, et~al.
\newblock Mastering chess and shogi by self-play with a general reinforcement
  learning algorithm.
\newblock \emph{arXiv preprint arXiv:1712.01815}, 2017.

\bibitem[Strehl \& Littman(2008)Strehl and Littman]{strehl2008analysis}
Alexander~L Strehl and Michael~L Littman.
\newblock An analysis of model-based interval estimation for markov decision
  processes.
\newblock \emph{Journal of Computer and System Sciences}, 74\penalty0
  (8):\penalty0 1309--1331, 2008.

\bibitem[Strehl et~al.(2009)Strehl, Li, and Littman]{strehl2009reinforcement}
Alexander~L Strehl, Lihong Li, and Michael~L Littman.
\newblock Reinforcement learning in finite mdps: Pac analysis.
\newblock \emph{Journal of Machine Learning Research}, 10\penalty0
  (Nov):\penalty0 2413--2444, 2009.

\bibitem[Sutton et~al.(2000)Sutton, McAllester, Singh, and
  Mansour]{sutton2000policy}
Richard~S Sutton, David~A McAllester, Satinder~P Singh, and Yishay Mansour.
\newblock Policy gradient methods for reinforcement learning with function
  approximation.
\newblock In \emph{Advances in neural information processing systems}, pp.\
  1057--1063, 2000.

\bibitem[Tang et~al.(2017)Tang, Houthooft, Foote, Stooke, Chen, Duan, Schulman,
  DeTurck, and Abbeel]{tang2017exploration}
Haoran Tang, Rein Houthooft, Davis Foote, Adam Stooke, OpenAI~Xi Chen, Yan
  Duan, John Schulman, Filip DeTurck, and Pieter Abbeel.
\newblock \# exploration: A study of count-based exploration for deep
  reinforcement learning.
\newblock In \emph{Advances in Neural Information Processing Systems}, pp.\
  2750--2759, 2017.

\bibitem[Thompson(1933)]{thompson1933likelihood}
William~R Thompson.
\newblock On the likelihood that one unknown probability exceeds another in
  view of the evidence of two samples.
\newblock \emph{Biometrika}, 25\penalty0 (3/4):\penalty0 285--294, 1933.

\bibitem[Tsividis et~al.(2017)Tsividis, Pouncy, Xu, Tenenbaum, and
  Gershman]{tsividis2017human}
Pedro~A Tsividis, Thomas Pouncy, Jacqueline~L Xu, Joshua~B Tenenbaum, and
  Samuel~J Gershman.
\newblock Human learning in atari.
\newblock 2017.

\bibitem[Weber et~al.(2017)Weber, Racani{\`e}re, Reichert, Buesing, Guez,
  Rezende, Badia, Vinyals, Heess, Li, et~al.]{weber2017imagination}
Th{\'e}ophane Weber, S{\'e}bastien Racani{\`e}re, David~P Reichert, Lars
  Buesing, Arthur Guez, Danilo~Jimenez Rezende, Adria~Puigdomenech Badia, Oriol
  Vinyals, Nicolas Heess, Yujia Li, et~al.
\newblock Imagination-augmented agents for deep reinforcement learning.
\newblock \emph{arXiv preprint arXiv:1707.06203}, 2017.

\end{thebibliography}
\bibliographystyle{iclr2019_conference}

\end{document}